\documentclass[conference]{IEEEtran}
\IEEEoverridecommandlockouts
\usepackage{cite}
\usepackage{amsmath,amssymb,amsfonts}
\usepackage{algorithmic}
\usepackage{graphicx}
\usepackage{textcomp}

\usepackage{subcaption}
\usepackage{graphicx} 

\usepackage{comment}

\newcommand{\bphi}{\bar \phi}
\newcommand{\0}{\mathbf {0}}

\usepackage{url}

\usepackage{xcolor}
\def\BibTeX{{\rm B\kern-.05em{\sc i\kern-.025em b}\kern-.08em
    T\kern-.1667em\lower.7ex\hbox{E}\kern-.125emX}}
\begin{document}

\title{Autonomous Learning of Attractors for Neuromorphic Computing with Wien Bridge Oscillator Networks\\
\thanks{This work was performed under the auspices of the U.S. Department of Energy (DOE)
at Los Alamos National Laboratory, operated by Triad National Security, LLC,
under Contract No.~89233218CNA000001, and was supported in part by the DOE Advanced
Scientific Computing Research (ASCR) program under Award No.~DE-SCL0000118.}
}

\author{%
\IEEEauthorblockN{%
Riley~Acker\IEEEauthorrefmark{1},
Aman~Desai\IEEEauthorrefmark{1}\IEEEauthorrefmark{2},
Garrett~Kenyon\IEEEauthorrefmark{1},
Frank~Barrows\IEEEauthorrefmark{2}\IEEEauthorrefmark{3}}
\IEEEauthorblockA{\IEEEauthorrefmark{1}Computing and Artificial Intelligence (CAI) Division, 
Los Alamos National Laboratory, Los Alamos, NM 87545, USA
}
\IEEEauthorblockA{\IEEEauthorrefmark{2}Center for Nonlinear Studies (CNLS), 
Los Alamos National Laboratory, Los Alamos, NM 87545, USA}
\IEEEauthorblockA{\IEEEauthorrefmark{3}Theoretical Division (T Division), 
Los Alamos National Laboratory, Los Alamos, NM 87545, USA\\
Email: fbarrows@lanl.gov}
}

\maketitle

\begin{abstract}
We present an oscillatory neuromorphic primitive implemented with networks of coupled Wien bridge oscillators and tunable resistive couplings. Phase relationships between oscillators encode patterns, and a local Hebbian learning rule continuously adapts the couplings, allowing learning and recall to emerge from the same ongoing analog dynamics rather than from separate training and inference phases. Using a Kuramoto-style phase model with an effective energy function, we show that learned phase patterns form attractor states and validate this behavior in simulation and hardware. We further realize a 2-4-2 architecture with a hidden layer of oscillators, whose bipartite visible-hidden coupling allows multiple internal configurations to produce the same visible phase states. When inputs are switched, transient spikes in energy followed by relaxation indicate how the network can reduce surprise by reshaping its energy landscape. These results support coupled oscillator circuits as a hardware platform for energy-based neuromorphic computing with autonomous, continuous learning.
\end{abstract}

\section{Introduction}
Neuromorphic computing has the potential to form the basis of next-generation computers by leveraging brain-inspired computational paradigms to natively implement brain-like information processing \cite{Kudithipudi2025,Indiveri_IEEE_2015}. Conventional von Neumann architectures separate memory from computation and require substantial energy and infrastructure to train and run large-scale models, whereas biological brains perform continuous learning and inference with modest power on noisy, analog, and massively parallel substrates \cite{Mead_2002}. The challenge, therefore, is to identify the relevant information-processing and computational functions of the brain that we hope to recapitulate in hardware and materials, and to design neuromorphic systems and functional devices that can directly embed these dynamics into their physical operation.

Much neuromorphic work focuses on the complexity of individual neurons or on spiking neural networks and their associated encoding schemes. Equally important, however, are the rich collective dynamics that emerge at the population level. Neural activity exhibits widespread oscillations and synchronization, and both experimental and theoretical studies suggest that relative phases and attractor dynamics play key roles in distributed memory and computation \cite{Buzsaki_Science_2004,Fries_CogSci_2005,Hopfield_PNAS_1982}. Kuramoto-type models provide a canonical description of coupled oscillators and phase synchronization \cite{kuramoto2003chemical}, while Hopfield networks provide a canonical model of distributed associative memory \cite{Hopfield_PNAS_1982}. Here we are interested in physically combining these paradigms, using coupled oscillators to realize phase-coded attractor states.

Physical neural networks can leverage the natural energy-minimization tendencies of physical systems. When the dynamics of a circuit 
approximate gradient descent on an energy function, attractor states correspond to local minima of that energy, and local learning rules can be framed as modifications of the underlying landscape. Such an energy function enables learning schemes such as Hebbian plasticity or equilibrium-propagation-like updates to be implemented directly in hardware, without explicit digital optimization loops \cite{Scellier_Frontiers_2017,lecun2006tutorial,Xie2003EquivalenceOB}. Beyond simply converging to fixed points, physically realized networks can operate in non-equilibrium steady states and transition between stable and near-critical regimes, potentially enriching their computational repertoire.


Our approach implements several core features of biological learning that are relevant to near-term neuromorphic hardware. Learning is autonomous, with synaptic updates driven solely from local phase interactions as opposed to an external clock. The network also learns continuously, as the Hebbian learning rule remains active during both learning and recall as opposed to being restricted to a dedicated training phase. Collectively, these properties support true local learning, as the update rule relies solely on information present at each individual synapse. 

Oscillatory circuits based on analog components, such as Wien bridge oscillators \cite{English_PRE_2015,English_PRE_2016}, provide a natural platform for such dynamics: they are simple to fabricate, exhibit robust limit cycles, and can be coupled through tunable resistive elements that play the role of synapses. 

In this work we present a neuromorphic primitive based on networks of coupled Wien bridge oscillators that implements phase-based associative memory with autonomous, continuous learning. Phase relationships between oscillators encode input and output patterns, and a local Hebbian learning rule with an Oja-inspired decay term continuously adapts coupling strengths during both ``learning'' and ``recall'', allowing the same physical dynamics to support storage and retrieval. We study both small feedforward architectures and a multilayer 2--4-2 architecture with a hidden layer of oscillators, whose bipartite coupling structure mirrors that of a  Boltzmann machine and gives rise to a non-unique internal energy landscape. 
While Kuramoto models with Hebbian plasticity have been analyzed previously in adaptive oscillatory networks \cite{Seliger_PRE_2002, Nishii_NeurNet_1998}, we are not aware of prior work combining our specific rule with a multi-layer Kuramoto phase-coupled neural network for continuous learning and recall \cite{NISHIKAWA_physicaD_2004,Nishikawa_PRL_2005,Du_JChemPhys_2024}.

\section{Circuits}

Widely used in early analog electronics and instrumentation, the Wien bridge oscillator  has several attractive properties for neuromorphic implementations.  
Each oscillator produces a stable sinusoidal oscillation, and the natural oscillation frequency is tunable and set by its series-parallel RC network. 
Resistive coupling between oscillators allows them to ``pull'' their phases together when connected to the non-inverting input, or “push” them apart when connected to the inverting input. 
Importantly, oscillations have a fixed voltage amplitude, making signal propagation, synchronization, and modeling possible. The Wien bridge oscillator is also easily prototyped, and thus serves as an analog to Kuramoto-style phase synchronization and learning.

\subsection{Wien Bridge Oscillator Network}
\begin{figure}[h] 
    \centering
    \includegraphics[width=0.95\linewidth]{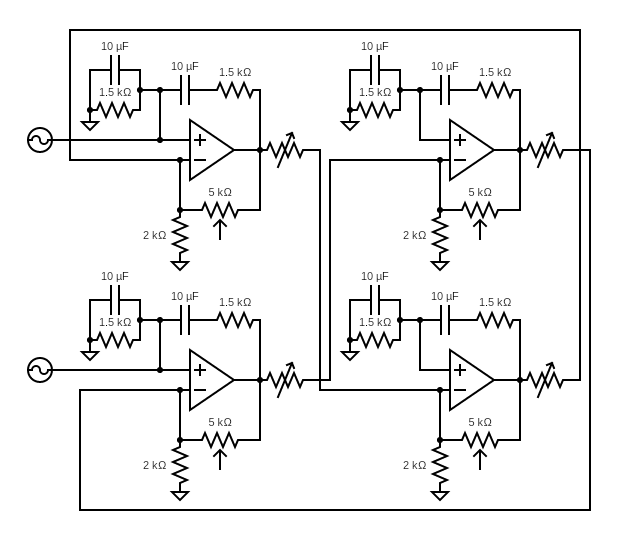}
    \caption{ Physical circuit implementation of a Hebbian weight matrix. Each neuron is implemented as a Wien bridge oscillator with intrinsic frequency set by its series–parallel RC feedback network; for matched components $R$ and $C$ in the bridge $f\approx (2\pi RC)^{-1}$.  Resistors between oscillators implement synaptic weights, determined by the Hopfield outer-product for the two encoded patterns, and potentiometers are used in the gain network to stabilize amplitude and minimize distortion. Analog sources inject the input pattern into the first two nodes, and all oscillator output voltages are measured using an oscilloscope. 
     }
    \label{fig:circuit}
\end{figure}

We construct a network of Wien bridge oscillators consisting of four oscillators, which act as neurons. 
While motivated by the oscillatory dynamics of neuronal populations, we refer to each oscillator as a ``neuron'' for consistency with the nomenclature of neural networks.
Each neuron is connected with variable 
resistors, implemented as 
digital potentiometers  (AD5242BRZ1M, Analog Devices Inc.), 
this is shown schematically in Figure \ref{fig:circuit}.

External drive signals on input neurons were generated using an AD9833 direct digital synthesis (DDS) module controlled by an \texttt{Arduino Nano Every} microcontroller to set the phase and frequency digitally, effectively clamping the input oscillators. The DDS output was passed through a non-inverting amplifier and connected to the input oscillator. The second input oscillator was either driven by a voltage follower from the first non-inverting amplifier or an inverter, depending on the desired phase relationship of the inputs. In scaled implementations, a multichannel digital-to-analog converter (DAC) would provide independent drive signals to each input oscillator. The feedback resistor network that sets the amplifier gain uses 3296W multi-turn trimmer potentiometers, which allow the amplifier gain to be finely adjusted to compensate for component tolerances. 
As the circuit does not include automatic gain control (AGC) elements, 
oscillations slightly clip at the supply rail when the operational amplifier output approaches its supply voltages. 
The Wien bridge oscillators are built from readily available 1\% metal–film resistors and 10\% multilayer ceramic capacitors. Circuits were implemented on a breadboard using TLV2461 operational amplifiers (op-amps) powered at $1.5$ V, and signals were measured with a \texttt{Siglent Technologies SDS1104X-E} four channel oscilloscope. The low voltage operating regime was chosen to maintain compatibility with typical memristive elements, whose operating ranges lie within this range. 

Circuit simulations were performed in LTSpice using idealized components matched to those used in physical prototypes, as well as the TLV261 op-amp macromodel. Voltage sources with voltage followers implemented the input biasing. Due to lack of parasitics and slight deviations from physical hardware, coupling resistances 
were increased to obtain distortion comparable to the measured circuit. Synaptic weights were implemented as resistive couplings: routing an oscillator’s output to the non-inverting input of another corresponds to positive coupling, while routing to the inverting input corresponds to negative coupling.
Circuit architecture is provided in Figure \ref{fig:circuit}.

\subsection{Characterization of Energy Function}

To train physical hardware, one often needs to find a suitable energy function; here we analyze our Wien bridge oscillator circuit in hardware and find that it was well-described by a Kuramoto energy function. 

The Kuramoto model is an established description of coupled oscillator networks. For a system of $N$ oscillators, where oscillator $i$ has time-dependent phase $\phi_i$, the Kuramoto dynamics of oscillator $i$ are given by the differential equation: \begin{align}
\label{kuramoto_dynamics}
    \dot{\phi_i} = \omega_i
    - \sum_{j}
    K_{i, j} \sin(\phi_i - \phi_j ),
\end{align}
where $\omega_i$ is the natural frequency of oscillator $i$ and $K_{ij}$ represents the coupling between oscillators $i$ and $j$. We separate the natural frequency into a global rotating reference frame $\omega_o$ and the local difference in the natural frequency, $\Delta \omega_i$, (generally due to dispersion of the natural frequency among devices) such that $\omega_i=\omega_o+\Delta\omega_i$. We define $s_i$ as the phase-state vector on the unit circle of oscillator $i$,  let the matrix $K$ be symmetric, and we have the corresponding energy function
\begin{equation}
    F(\phi)=-\sum_{i<j} K_{ij}s_i\cdot s_j -\sum_i (\omega_o+\Delta\omega_i) s_i
\label{eq:Hopefield_w-bias}
\end{equation}
which is a standard Hopfield energy. In other words, the Kuramoto energy function is the natural generalization of the Hopfield energy to oscillating states. We work in a rotating frame, wherein we set $w_o\equiv 0$. 
The Kuramoto model with homogeneous natural frequency in the rotating frame transforms to the Hopfield model, 
\begin{equation}
    F(\phi)=-\sum_{i< j} K_{i,j}s_i\cdot s_j  .
\label{eqn:HopfieldEnergy}
\end{equation}
Thus, with the right data encoding we can get our network to act as a Hopfield network.

\begin{figure}[h] 
    \begin{subfigure}{0.25\textwidth}
        \includegraphics[width=1\linewidth]{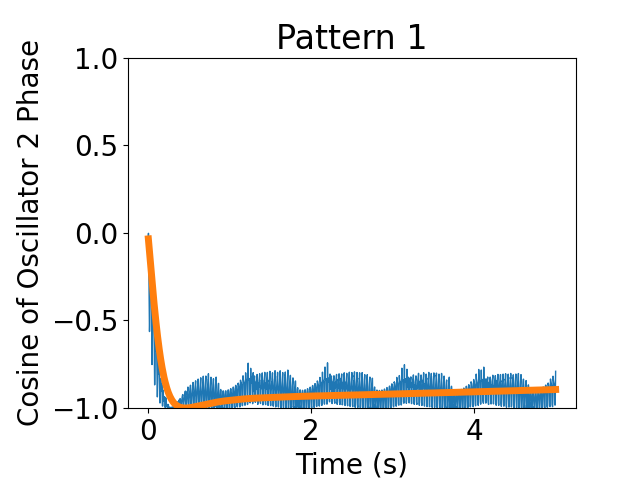}
    \end{subfigure}%
    \begin{subfigure}{0.25\textwidth}
        \includegraphics[width=1\linewidth]{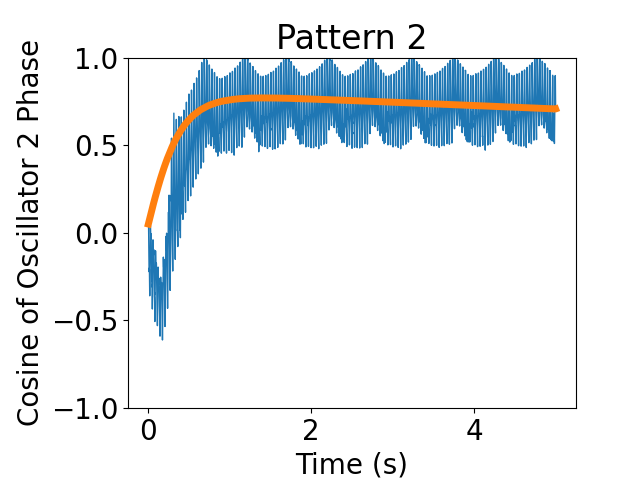}
    \end{subfigure}%
    \caption{ Plots of the cosine phase, $\cos(\phi)$, of an unbiased neuron in the $2 \times 2$ oscillator circuit (Figure \ref{fig:circuit}) are shown in blue, dynamics simulated by LTspice. The corresponding cosine phase determined by the fit Kuramoto model dynamics are shown in orange. Pattern 1 and pattern 2 correspond to distinct phases on the biased neurons, phase is set by the driving bias; with  two oscillators driven at phases of $(0,\pi)$ and $(0,0)$ in pattern 1 and pattern 2, respectively.      
    }
    \label{fig:fitted_kuramoto_dynamics}
\end{figure}

We fit the sinusoidal voltage time-series from LTspice simulations to the Kuramoto model. Time-dependent phase is determined using the extended Hilbert transform  \cite{Matsuki2023}. Coupling strengths and oscillator natural frequencies were estimated by fitting an idealized Kuramoto model to the measured phase trajectories, using gradient-based optimization through a numerical ODE solver implemented in JAX \cite{jax2018github, diffrax_citation, equinox_citation, optimistix_citation, hairer2008solving, tsitouras2011runge}. Figure  \ref{fig:fitted_kuramoto_dynamics} displays the cosine of the phase of an output oscillator as determined in LTSpice (blue line) and by the fit Kuramoto model (orange line). The hardware dynamics are described by the Kuramoto model, as demonstrated by the close correspondence between the mean of the hardware phase and the Kuramoto model dynamics. We note that the idealized Kuramoto dynamics seem to differ from the true time-averaged dynamics by a slow linear drift. We suspect that this deviation is due to fitting to the data which has a natural beat frequency due to incorporating hardware tolerances and the 'start-up' dynamics.

Importantly, the fitted Kuramoto dynamics accurately predict the phase difference at which the phases lock. This serves as empirical verification that the phase dynamics of coupled Wien bridge oscillators minimize the energy function given by eq. \eqref{eqn:HopfieldEnergy}. 

\section{Distributed Memory and Hebbian Learning}

We demonstrate training patterns in a small oscillator network. Here a pattern is the bit string we intend to encode, represented as a vector of four oscillator phases. Patterns and phases can be interpreted as logical binary data corresponding to phase of $0$ and $\pi$ relative to the rotating frame; we refer to this as a binarization of our phase,  $s_\mathrm{binary}$. While logical outputs are restricted to phases of $0$ and $\pi$ in the systems studied here, it is possible to generalize this approach to richer bit encoding schemes.  

Phases are transformed using $s_\mathrm{binary} = \cos(\phi)$, which yields $+1$ for $\phi=0$ and $-1$ for $\phi=\pi$; patterns are represented as a four-dimensional vector with entries in $\{\pm 1\}$. The attractor basin is determined by the outer-product of the $m$-stored patterns, $K=\frac{1}{m}\sum^m_i s^i_\mathrm{binary}\wedge s^i_\mathrm{binary}$ as in the standard Hopfield network. The diagonal is zeroed to prevent self-coupling and all components of the resulting matrix are then clipped to $[-1,1]$ for consistency with a dynamically stable and physically realistic range.

\subsection{Hebbian Learning}
To encode patterns we train the network by exposing it to the set of desired patterns. The two oscillators attached to the signal generators in Figure \ref{fig:circuit}  are \emph{input} neurons, with their phase controlled 
by the biasing. The two other \emph{output }neurons are nudged during training; they receive a 
nudging signal that drives the output phase towards the desired phase. This can be implemented with a small bias such that the phase of the output neuron, $\phi_O$ is driven towards the desired phase, $\phi^*_O$, implemented as 
\begin{equation}
\Delta {\phi^\mathrm{nudge}_\mathrm{output}}= - K_\mathrm{nudge}(\phi_\mathrm{output}-\phi^*_\mathrm{output}).
\end{equation}

Throughout learning, we employ a Hebbian learning rule to adjust the coupling strengths between oscillators, with changes in synaptic conductivity depending on the instantaneous phase relationships between oscillators. The cosine of the phase difference, $\phi_{i }- \phi_{j}$, is maximized when oscillators are in-phase and minimized when oscillators are anti-phase, following the canonical \emph{“neurons that wire together fire together”} principle of neuroscience. 
Simulated oscillators are assigned random initial phases, consistent with physical Wien bridge oscillators which naturally form oscillations from noise.

The weight update follows 
 the standard update procedure for energy functions, defined by the energy function $F(\phi)$ in eq. \eqref{eqn:HopfieldEnergy}. We implement weight updates using a weight decay term, consistent with biological synaptic normalization \cite{Oja1982}, which is motivated by a regularized energy function, $F_\mathrm{reg}(\phi):= F(\phi ) +\tfrac{\lambda}{2} \sum_{ij}K_{ij}^2$. This yields the continuous time and  regularized weight updates
 \begin{subequations}
\begin{align}
\dot{K}_{ij} &:=- \eta \partial_{K_{ij}}F(\phi)= \eta \cos(\phi_i-\phi_j ), \label{eq:HebbianUpdate_1} \\
\Delta {K}_{ij} &:= - \eta \partial_{K_{ij}}F_\mathrm{reg}(\phi) = \eta \left(\cos(\phi_i-\phi_j) -\lambda K_{ij}\right) .
\label{eq:HebbianUpdate_2}
\end{align}
\end{subequations}
The first term in eq. \eqref{eq:HebbianUpdate_2} is a Hebbian update based on local phase, while the $-\lambda K_{ij}$ term implements weight normalization that keeps the effective norm of $K$ bounded.

\begin{figure}[h] 
    \centering
   \includegraphics[width=1\linewidth]{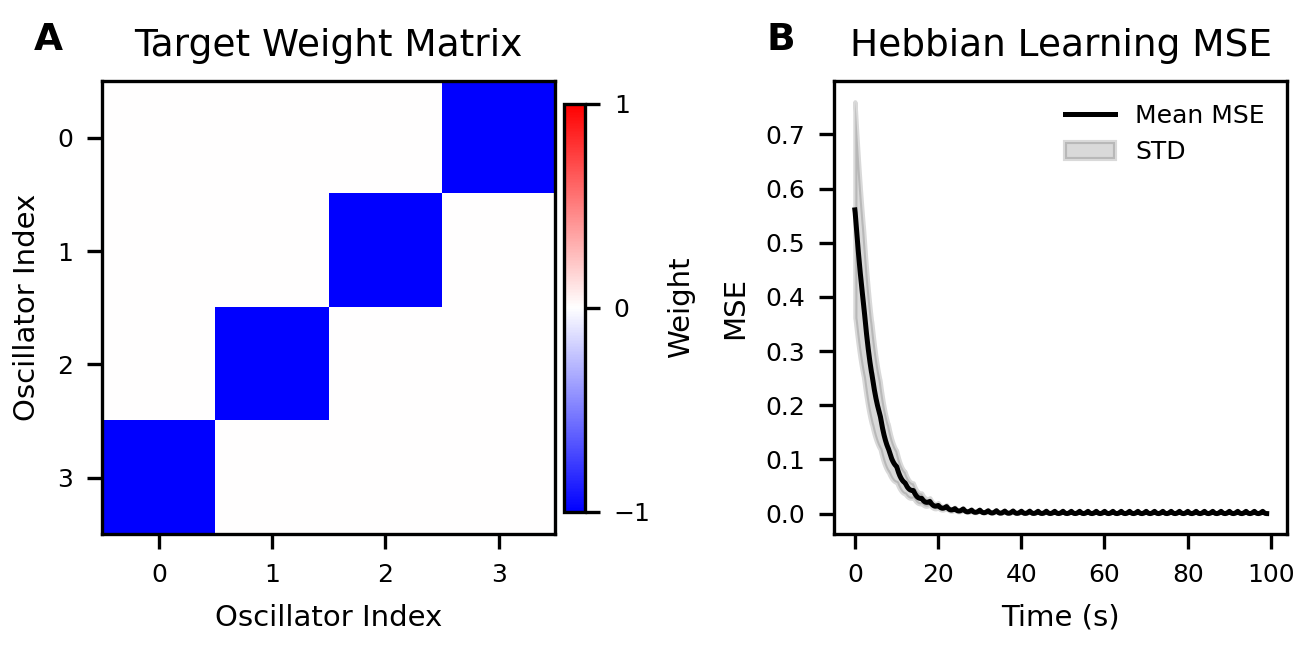}
    \caption{ 
    Hebbian learning in a $2\times 2$ network. A) The ideal weight matrix for the two desired patterns is obtained using the summed outer products of their $\pm 1$ representations; this matrix is used in numerical, SPICE and physical circuit implementations. B)  Average learning curve over 100 numerically simulated runs, with individual runs using a different pair of randomized patterns. A continuous Hebbian learning rule with a decay term is applied to a randomly initialized weight matrix. During training, input oscillators (0,1) are clamped and alternate between two desired patterns, while output oscillators (2,3) are nudged towards the correct phase relative to the inputs. At each time step, the MSE is computed by taking the element-wise difference between the current weight matrix and the ideal weight matrix, squaring the differences, and averaging them into a single scalar. Shaded region displays the standard deviation across runs.  }
    \label{fig:HopefieldTraining}
\end{figure}

These dynamics are simulated numerically using an Euler step approximation, whereas analog hardware can solve this dynamical equation in real time through physical coupling interactions. The target connectivity matrix and the error during training are shown in Figure \ref{fig:HopefieldTraining}.

Initial weights are drawn from a uniform distribution $[-0.25,0.25]$. 
At each Euler step, both the Kuramoto phase dynamics and the local Hebbian learning rule are applied to each oscillator. To train both patterns and their corresponding attractor landscape, the imposed pattern driving the network is alternated at fixed intervals between pattern 1, $[1,-1,1,-1]$, and pattern 2, $[1,1,-1,-1]$. This is implemented by clamping the input neurons to a given phase offset, enforcing a $0$ or $\pi$ radian difference. 
For the $2\times 2$ connected four neuron architecture, the mean squared error (MSE), Figure \ref{fig:HopefieldTraining} (b), is computed as the mean of the element wise squared differences between the current learned and ideal target weight matrix. In each run, the target matrix is calculated from the summed outer product of its pattern set.

\begin{figure}[h] 
    \centering
    \includegraphics[width=0.95\linewidth]{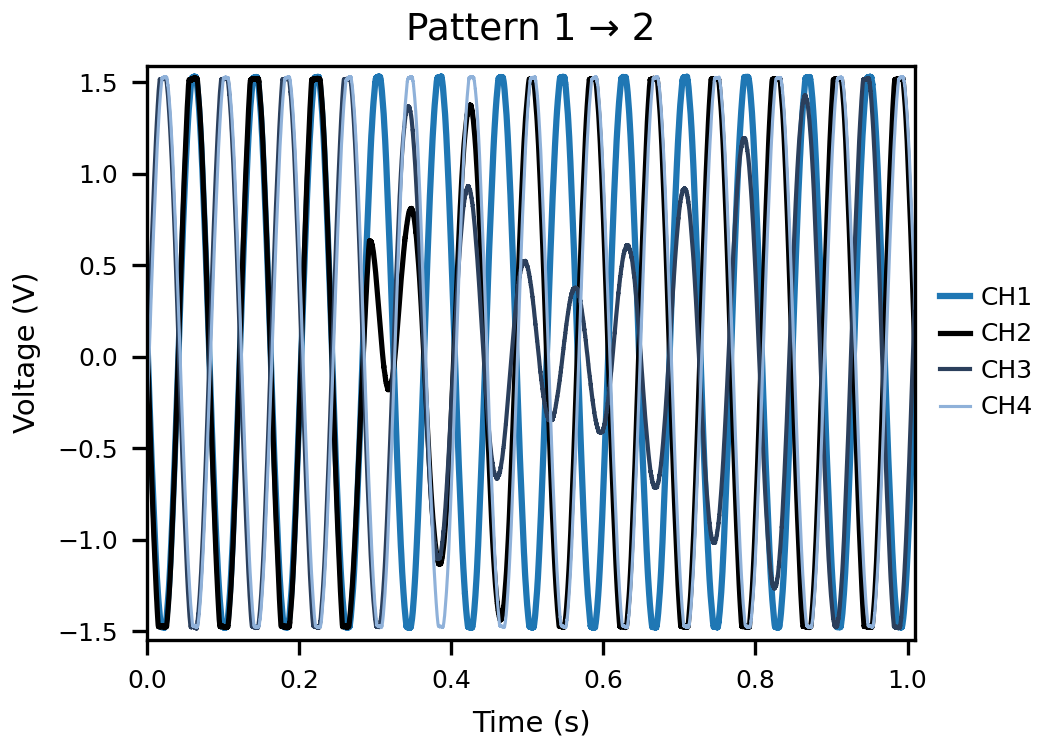}
    \caption{ Oscilloscope voltage measurements showing transitions between stored patterns: the circuit is driven to the first stored pattern and subsequently switched to the second, and the resulting waveforms capture the transient switching behavior and convergence to the corresponding attractor defined by Hopfield weight matrix. After switching, both patterns settle into stable steady states.
     }
    \label{fig:PatternTransitions}
\end{figure}

The circuit with the target weight matrix, e.g., Figure \ref{fig:HopefieldTraining} (a), was implemented in the physical circuit, shown in Figure \ref{fig:circuit}. 
Running this circuit then amounts to inference, or \emph{recall}, on the trained network; Figure \ref{fig:PatternTransitions} displays the switching of the two attractors, 
and the resulting trajectories 
demonstrate reliability of the dynamics when instantiated in hardware. 

We initialize the run with an input bias corresponding to one of the two trained patterns. We observe phase locking corresponding to the trained pattern, demonstrating associative memory.  Part way through the simulation, $\sim 0.3$ seconds, the input bias is switched to the other trained pattern. We observe the output neurons (channel 3 and channel 4) transiently adjust their oscillations and converge to a phase locked configuration corresponding to the other trained pattern; thus demonstrating switching attractors.

\begin{figure}[h] 
    \centering
    \includegraphics[width=1\linewidth]{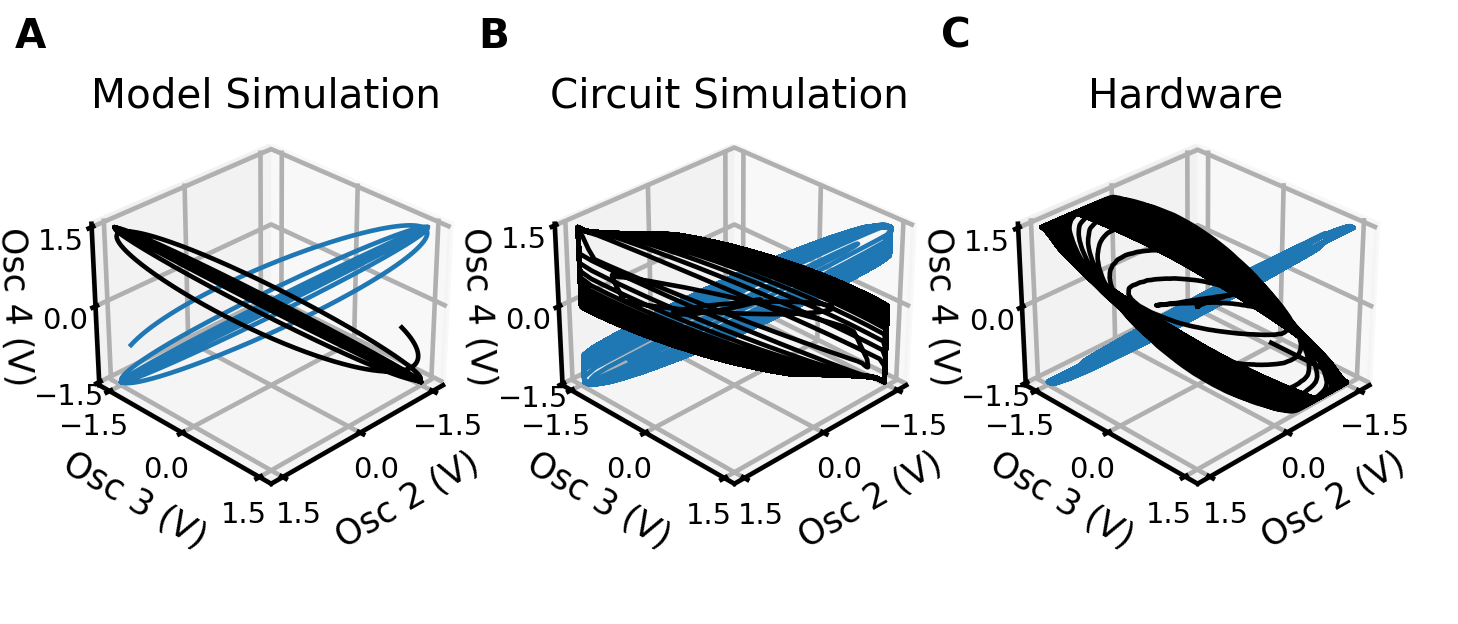}
    \caption{ Trajectories of oscillators in model simulations, SPICE, and hardware. Pattern 1 (blue) encoded as [1,-1,1,-1], and pattern 2 (black) as [1,1,-1,-1]. A)  A $2\times 2$ oscillator architecture is simulated numerically using ideal Kuramoto dynamics. Voltage trajectories are plotted from startup with color representing the encoded pattern. B) The same architecture is implemented in SPICE using Wien bridge oscillators, with 2\% capacitor and 1\% resistor tolerances sampled uniformly for the RC frequency-setting network of each oscillator. C) The same Wien bridge oscillator circuit is implemented in hardware, including 10\% capacitors and 1\% resistors; smoothed voltage traces are plotted from oscilloscope measurements. 
     }
    \label{fig:HopefieldInference}
\end{figure}

In Figure \ref{fig:HopefieldInference}, inference is demonstrated as two distinct attractors in Kuramoto model-level simulations (Python), circuit-level simulations (SPICE), and hardware measurements. We implement the same circuit configuration in Figure \ref{fig:circuit} and the weight matrix from Figure \ref{fig:HopefieldTraining} (a). Each pattern corresponds to a distinct attractor, indicated with blue and black lines. 

Here we see good correspondence between the trajectories in model-level simulations, circuit-level simulations, and hardware measurements.  In the ideal Kuramoto model (a) the attractor converges towards a straight line, while simulated part tolerances (b), and true hardware imperfections (c) introduce deviations from the idealized trajectory.  Differences in the attractors can be attributed 
to hardware non-idealities, including how the oscillations clip due to op-amp transistor saturation, as well as parasitics and hardware tolerances.

\section{Continuous Autonomous Learning with Hidden Neurons}

\begin{figure}[h] 
    \centering
    \includegraphics[width=0.8\linewidth]{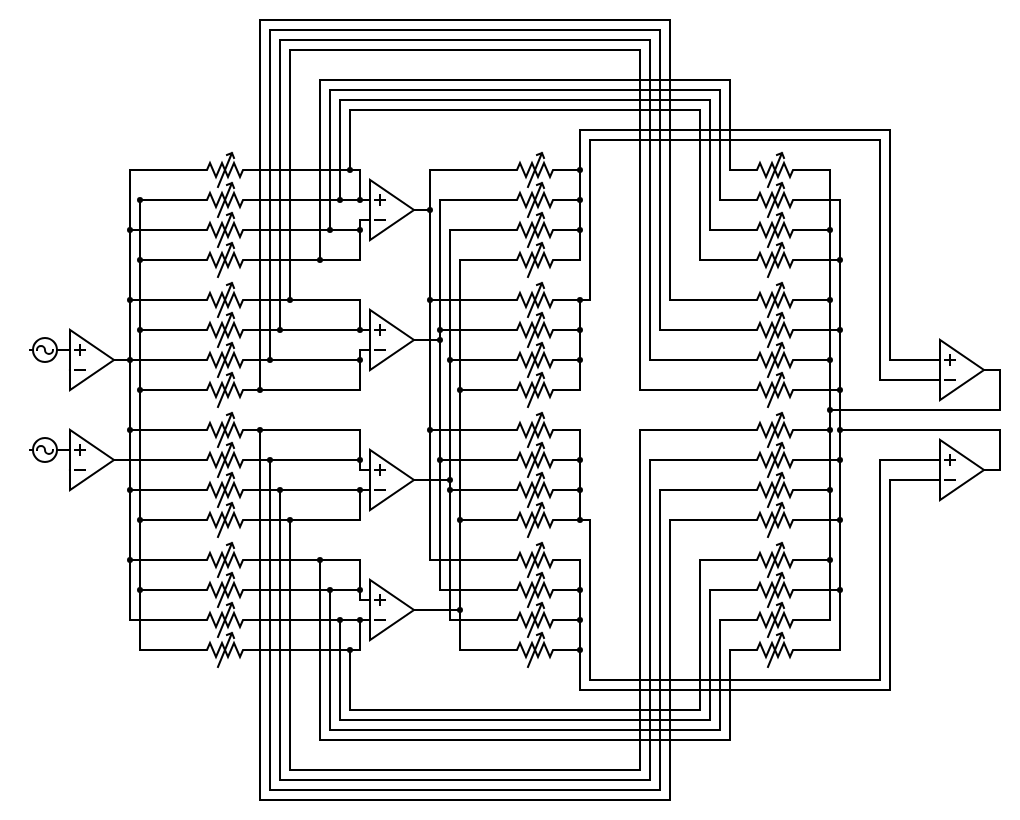}
    \caption{  In a physical 2–4–2 implementation, weights are realized with variable resistors. Each synapse uses a dedicated resistor feeding both the non-inverting and inverting inputs, and Hebbian learning drives the bidirectional weights in the hidden–output layer towards symmetric coupling.}
    \label{fig:2-4-2}
\end{figure}

We extend the 4-oscillator network to a $2$–$4$–$2$ architecture, consisting of two input oscillators, four hidden oscillators, and two output oscillators, interacting through learned Kuramoto couplings. 
This platform enables richer attractor dynamics, and is shown schematically in Figure \ref{fig:2-4-2}; for clarity, the RC frequency setting and gain networks are omitted but correspond to those shown in Figure \ref{fig:circuit}.

In Kuramoto model simulations, the same approach established in eq. \eqref{eq:HebbianUpdate_2} is used, with the decay term and Hebbian learning rate tuned for the layered configuration. Here, a continuous learning rule is used, and synaptic updates occur during both the “learning” and “recall” phases. 

During learning, input neurons are driven at a specific phase, corresponding to their respective binary value, and similarly the phase on the output neurons are nudged towards their respective binary value. 
Training consists of alternating the two patterns and allowing the weights to update continuously. 
During recall, i.e., inference, the external drive on the output neurons is removed, allowing the attractor dynamics to unfold naturally in accordance with the weight matrix.

\begin{figure}[h] 
    \centering
    \includegraphics[width=0.95\linewidth]{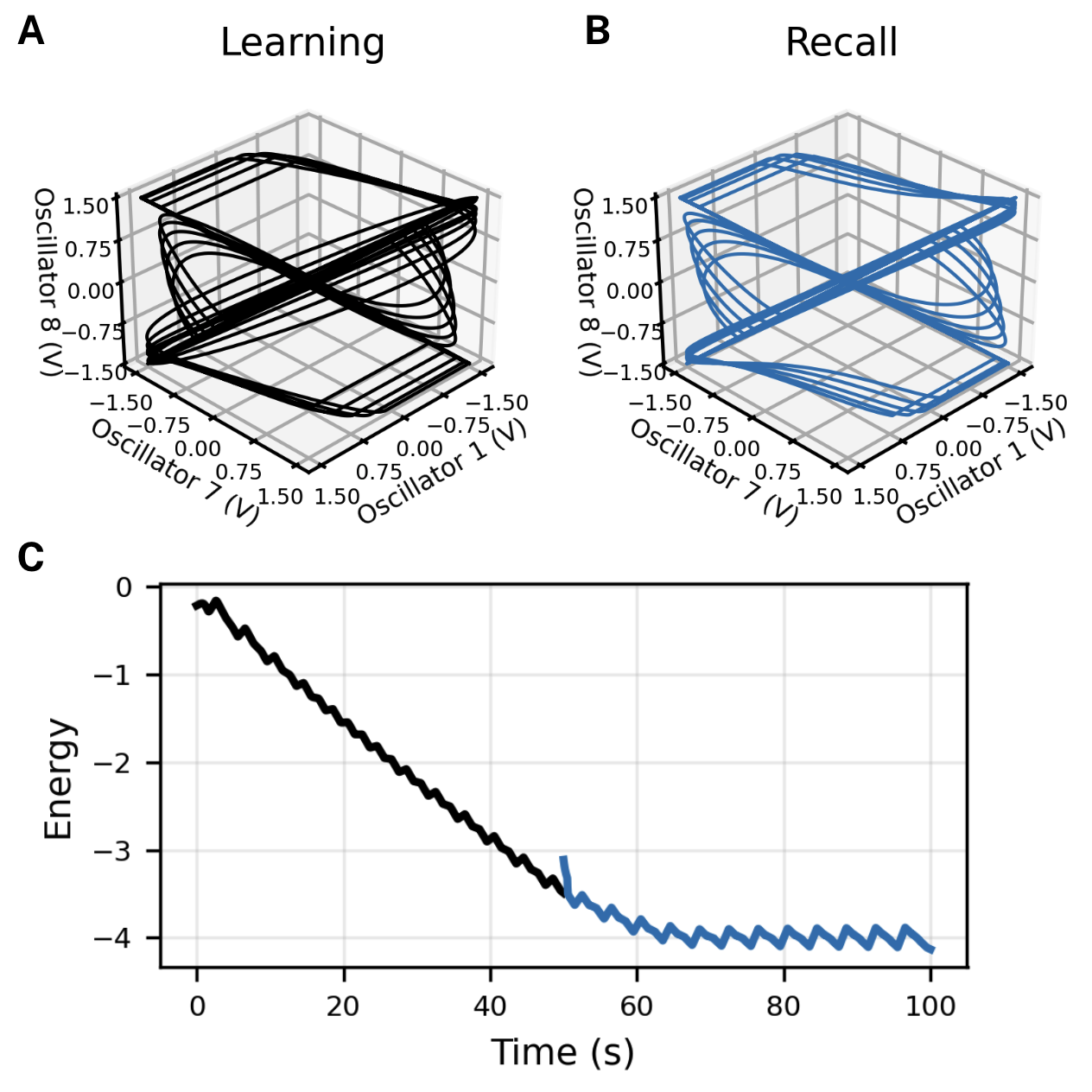}
    \caption{    Continuous Hebbian learning in a three-layer oscillatory neural network. A) \textit{Learning:} Kuramoto dynamics are simulated numerically and weight updates are applied at each timestep using a Hebbian learning rule with a decay term. Input oscillators are clamped to  voltage sources encoding the desired patterns, and output oscillators are nudged toward the correct phase relationships. After a settling period, oscillator voltages are plotted in 3D to illustrate the emerging attractor dynamics.  B) \textit{Recall:} Oscillator phases are initialized randomly and output oscillators evolve freely under the learned weights. Inputs alternate between stored patterns, and resulting trajectories are plotted. 
    C) A single scalar Hopfield energy value is computed from the cosine of all hidden-input phase differences, illustrating overall alignment of oscillator phases and that energy decreases both as the network learns and settles into an attractor. Energy values are smoothed with a centered windowed average and plotted over time, with the black line indicating training and blue indicating recall.
    }
    \label{fig:LearningRecall}
\end{figure}

In Figure \ref{fig:LearningRecall},  oscillator voltages are plotted from the two outputs (oscillators 7 and 8) and one input (oscillator 1), during learning (a) and recall (b). We observe two distinct dominant attractors, evident as the denser regions of parameter space, corresponding to the two alternating patterns. Learning and recall trace out similar trajectories. 
In Figure \ref{fig:LearningRecall} (c), Hopfield energy according to  eq. \eqref{eqn:HopfieldEnergy} is plotted during training (black) and recall (blue); throughout training, the energy decreases while bidirectional weights converge towards symmetry. Energy displays a sawtooth pattern during both training and recall corresponding to switching between patterns, and remains stable during inference. Such a sudden increase in energy is a \emph{surprise} which the network minimizes by switching attractor state. 
If we model phase configurations by a Gibbs distribution $p(\phi)\propto e^{-\beta F (\phi)}$, then the information–theoretic surprise of a configuration $\phi^*$ is
\begin{equation}
-\ln p(\phi^*) = \beta E(\phi^*) + \ln Z,
\end{equation}
with $Z$ the partition function. Thus, for fixed temperature $\beta>0$, surprise is a monotonic function of the energy, and the abrupt jumps we observe in $E(\phi(t))$ during pattern switches are exactly the corresponding jumps in surprise.

\begin{figure}[h] 
    \centering
    \includegraphics[width=0.9\linewidth]{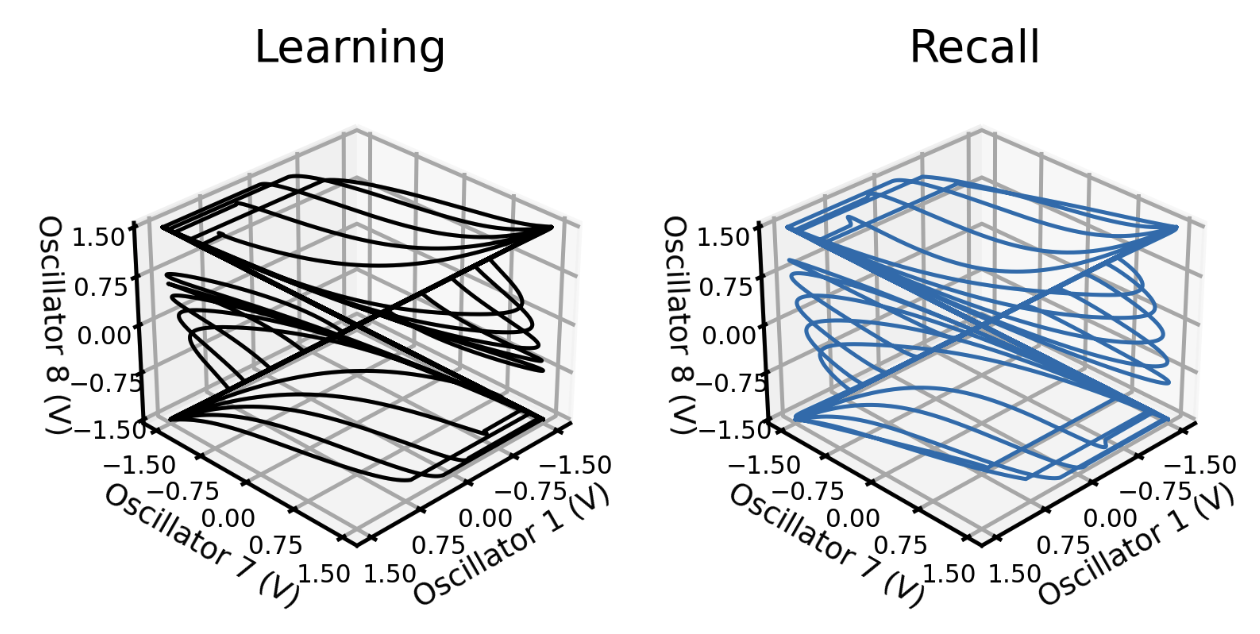}
    \caption{Learning and recall trajectories from training a $2-4-2$ network with a randomly initialized weight matrix.} 
    \label{fig:LearningRecall_example2}
\end{figure}

In Figure \ref{fig:LearningRecall_example2}, trajectories during learning and training  are shown for a 
separate run for the same learning task, with trajectories differing from those in Figure \ref{fig:LearningRecall} due to being initialized with randomized phases, weights, and natural frequency dispersion. 
In model simulations with differing patterns, recall trajectories closely trace out the learned trajectories, and are tolerant of frequency dispersion up to $20\%$ across oscillators. 

\begin{figure}[h] 
    \centering
    \includegraphics[width=0.85\linewidth]{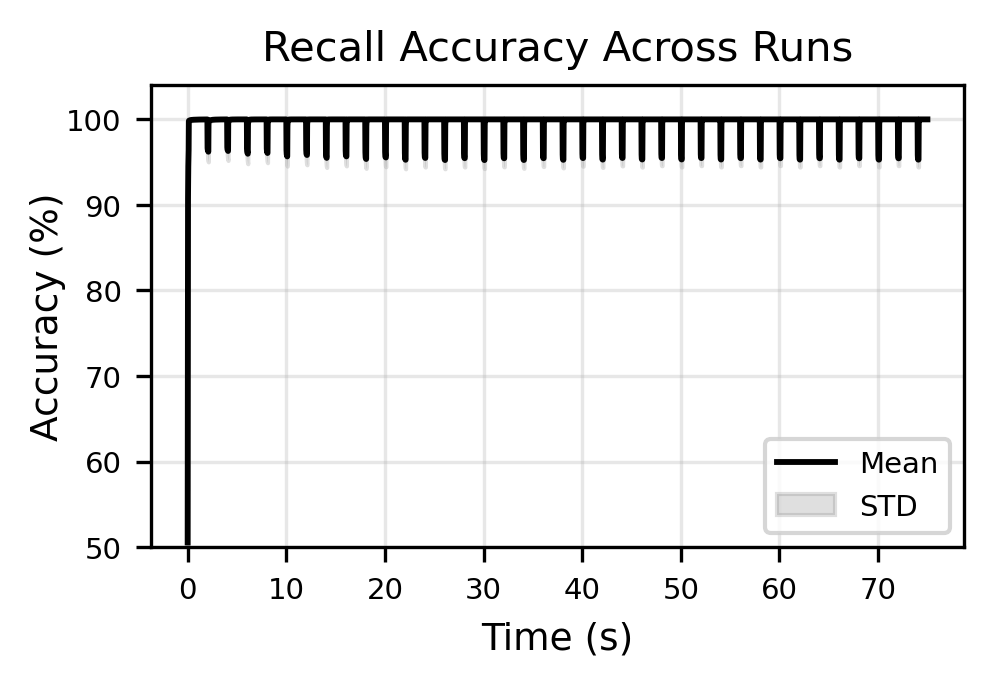}
    \caption{Accuracy of 2-4-2 architecture implementing continuous learning. The mean recall accuracy across 100 independent simulated runs plotted with a 100-timestep rolling average; input patterns are switched every $2.0$ seconds. 
    Each run first learns two randomly generated input-output patterns, with output oscillators released from external drive during recall. The recall accuracy of each output oscillator is computed in the rotating frame as the cosine similarity between observed and desired phase, $\phi$ and $\phi^*$, respectively; $\mathrm{Acc}= \tfrac{1}{2 } (\cos( \phi^* -\phi)+1)$. 
    The desired phase alignment and anti-phase error correspond to an accuracy of $100\%$  and $0\%$. Transient drops in accuracy occur when clamped inputs alternate between imposed patterns, with an initial accuracy at $t=0\,s$  reflecting randomly initialized phases. 
}
    \label{fig:Recall_Runs}
    \end{figure}

During recall, performance is assessed by alternating between stored input patterns and verifying whether the system reliably switches to the corresponding learned attractor.  
Figure \ref{fig:Recall_Runs} displays the recall accuracy, demonstrating the circuit does successfully reproduce the trained input patterns. Spikes in the accuracy plot correspond to transient errors when the input bias switches between two trained input patterns.  
    


\section{Stability}

Essential to this analysis is that the dynamics of the oscillators are robust to noise and perturbations, such as frequency dispersion. The dispersion of the natural frequencies acts as a local bias, as seen in eq. \eqref{eq:Hopefield_w-bias}. 

We have the canonical time evolution under an energy function given by
\begin{align}
    \dot \phi
    =
    -{\partial  F \over \partial \phi} .
\end{align}
We define $\phi^*_{orig}$ as a local minimum of the energy function $F(\phi)$ as formulated in \eqref{eqn:HopfieldEnergy}, and assume that $\nabla^2_{\phi} F(\phi^*_{orig})$ is invertible. If $\phi^*_{new}$ represents the new fixed point after frequency dispersion is added, then by the implicit function theorem,
\begin{align}
    \phi^*_{new} &\approx \phi^*_{orig} - \left (\nabla^2_{\phi} F(\phi^*_{orig}) \right )^{-1} \left ( \frac{\partial^2 F(\phi^*_{orig})}{ \partial \phi \partial \Delta \omega}  \right) \Delta \omega
    \nonumber \\
    &\approx 
    \phi^*_{orig} + \left (\nabla^2_{\phi} F(\phi^*_{orig}) \right )^{-1} \Delta \omega,
\end{align}
where the last equality uses the fact that the matrix of cross-derivatives is the negative identity, $-I$.

Note, the spectral norm of the inverse Hessian satisfies $\| \nabla^2_{\phi} F(\phi^*_{orig})^{-1} \|_2 = {1 \over \lambda_{min}(\nabla^2_{\phi} F(\phi^*_{orig}))} $, so the magnitude of the perturbation upon incorporating frequency dispersion is bound as
\begin{align}
    \| \phi^*_{orig} -  \phi^*_{new}\|_2
    &\approx
    \| \left (\nabla^2_{\phi} F(\phi^*_{orig}) \right )^{-1} \Delta \omega \|_2
   \nonumber  \\
    & \leq
    {\| \Delta \omega\|_2 \over \lambda_{min}(\nabla^2_{\phi} F(\phi^*_{orig}))}. 
\end{align}
Thus, $\lambda_{min}(\nabla^2_{\phi} F(\phi))$ can be viewed as a measure of the sensitivity of our architecture to frequency dispersion. When $F$ is more ``bowl-shaped'' around $\phi^*$, corresponding to larger $\lambda_{min}(\nabla^2_{\phi} F(\phi))$, the effect of frequency dispersion is reduced.


\section{Nonconvex learning}
This theoretical analysis emphasizes the fact that hidden layers add significant complexity to the optimization problem of learning a correct coupling matrix, especially in contrast to the explicit outer product formulation for the coupling matrix given when there are no hidden units.
Consider the four–oscillator network with binary phases $s \in\{\pm 1\}^4$, (for simplicity, here we neglect the $_\mathrm{binary}$ subscript)  and energy $E=F(\phi)\vert_{\phi}$ determined by eq. \eqref{eqn:HopfieldEnergy}. We fix a target pattern, $s^*$,  one of a set $\{ s^\mu\}$ of $m$-stored patterns, with $m$ less than the storage capacity of the Hopfield network, and determine couplings by the scaled outer product $   K(\lambda) = \lambda \frac{1}{m}\sum_{s^i\in \{ s^\mu\}} s^i\wedge s^i$, $\lambda>0$, with the diagonal set to zero.  Then the energy for the trained states $s^*\in\{s^\mu \}$ and the untrained states $s\notin \{s^\mu \}$ can be written as $E_{K(\lambda)}(s^*)$ and $ E_{K(\lambda)}(s)$; the trained states are low-energy states. 
If we view the Gibbs distribution $  p_K(s) \propto \exp\bigl(-\beta E_K(s)\bigr) $ parametrized by the off–diagonal entries of $K$, then this is a regular exponential family for finite $\beta>0$. The Fisher information
\begin{equation}
   I(K) =   \mathbb{E}_{p_K}
  \bigl[    \nabla_K \log p_K(s)\,     \nabla_K \log p_K(s)^\top \Bigr]
\end{equation}
is  positive definite for generic $K$, so the model is regular for the visible spins.  In particular, once the target patterns $\{ s^\mu\}$ and a scalar $\lambda>0$ are chosen, the corresponding coupling matrix $K(\lambda)$ is fixed in closed form and uniquely determined by this choice, with no ambiguity in the couplings.

Now we introduce a hidden layer and let $v\in\{\pm1\}^4$ denote the visible oscillators (two inputs and two outputs) and $h\in\{\pm1\}^4$ the hidden oscillators in a $2$–$4$–$2$ architecture.  Take an energy
\begin{equation}
  E_\theta(v,h) = - v^\top W h - h^\top A h,
\end{equation}
where $\theta=(W,A)$ collects all couplings allowed by the $2$–$4$–$2$ graph; in our case we set $A$ to a zero-matrix as we do not implement coupling within the hidden layer. The visible distribution is the marginal
\begin{equation}
  p_\theta(v)
  \;=\;
  \frac{1}{Z(\theta)}
  \sum_{h} \exp\bigl(-\beta E_\theta(v,h)\bigr),
\end{equation}
with the partition function, $Z(\theta) =   \sum_{v,h} \exp\bigl(-\beta E_\theta(v,h)\bigr)$. Requiring a given visible pattern $v^*$ (corresponding to the desired input–output phase pattern) to be in one of the visible low-energy states imposes a set of nonlinear inequalities in $\theta$.  These constraints do not single out a unique parameter, there are infinitely many $\theta$ that yield the same $p_\theta(v)$ (and hence the same visible low-energy states).  In particular, the hidden layer provides a redundant internal representation, because the couplings are continuous and the hidden units are unobserved, there exist distinct parameter values $\theta$ that induce exactly the same visible Gibbs distribution $p_\theta(v)$, so
\begin{equation}
  [\theta]
   =
  \bigl\{\theta' : p_{\theta'}(v) = p_\theta(v)\ \ \forall v\bigr\}
\end{equation}
is a nontrivial equivalence class in parameter space.  
Thus the parameterization is globally non-identifiable. Since the visible distribution over configurations has fewer independent degrees of freedom than $\theta$ the continuous parameters, the map $\theta \mapsto \{\log p_\theta(v)\}_v$ necessarily has a nontrivial kernel, so there exist nonzero directions $u$ along which all log–probabilities are stationary:
\begin{equation}
  \frac{\partial}{\partial\varepsilon}\Big|_{\varepsilon=0}
  \log p_{\theta+\varepsilon u}(v) \;=\; 0
  \quad\text{for all } v,
\end{equation}
which implies that the Fisher information matrix
\begin{equation}
  I(\theta)
  \;=\;
  \mathbb{E}_{p_\theta}
  \bigl[
    \nabla_\theta \log p_\theta(v)\,
    \nabla_\theta \log p_\theta(v)^\top
  \bigr]
\end{equation}
satisfies $I(\theta)u = 0$.  
Thus, realizing the desired visible pattern requires solving a nonconvex  optimization problem over $\theta$. The training dynamics of the physical network are  selecting one representative from the singular equivalence class $[\theta]$ that implements the same visible behavior and learning is a singular inverse problem.

\section{Conclusion}
\label{sec:conclusion}
We have demonstrated that small networks of coupled Wien bridge oscillators can implement a continuous-time oscillatory neuromorphic primitive for associative memory. Phase relationships between oscillators encode input and output patterns, and a local Hebbian learning rule with decay shapes the effective coupling matrix so that desired phase patterns become attractors of the dynamics. The same rule operates during both ``learning'' and ``recall'', so learning and computation proceed autonomously in a single continuous dynamical process rather than in separate algorithmic phases.

Viewed through an energy-based lens, the network learns to minimize an effective energy over phase configurations. When the inputs are switched, we observe transient spikes in this energy before the system relaxes into a new attractor. For a Gibbs-like interpretation of the dynamics, these spikes correspond to brief increases in surprise, followed by a reduction as the network adapts. This association between phase-energy and surprise suggests a hardware-level mechanism for learning: local weight updates driven by ongoing dynamics can steer the system toward lower-energy, more predictable configurations without requiring an explicit external optimization loop. Although we do not attempt a full predictive-coding or free-energy formalization, this provides a useful bridge between the energy landscape of the hardware and familiar notions of prediction error in computational neuroscience.

A key practical observation is that this continuous learning rule is very tolerant of frequency dispersion across oscillators, a fundamental feature of real Wien bridge circuits with mismatched RC components. This suggests that the learning rule can compensate for fabrication variability and drift, making oscillatory neuromorphic systems more robust than purely static-design approaches.  

In the extended 2-4-2 architecture with a hidden layer, the bipartite coupling between visible and hidden oscillators mirrors that of a Boltzmann machine, and multiple internal weight configurations can realize effectively equivalent visible mappings. The hardware thus selects particular low-energy internal states from a non-unique energy landscape, illustrating both the expressive power and inherent non-uniqueness of hidden-layer oscillator networks. 

These results support coupled oscillator circuits as a promising architecture for energy-based neuromorphic computing. They combine continuous-time dynamics, local learning rules, robustness to device mismatch, and the ability to exploit hidden units for rich internal representations. 
Future work will scale these ideas to larger networks and richer pattern families, deepen quantitative links between hardware dynamics and probabilistic models, and incorporate memristive coupling elements to enable on-chip continuous Hebbian learning.
\section*{Acknowledgment}
We thank Siddharth Mansingh for helpful discussions and Aiping Chen for experimental support at the Center for Integrated Nanotechnologies (CINT).  
AD and FB also gratefully acknowledge support from the Center for Nonlinear Studies (CNLS) at LANL.
RA and GK acknowledge support from the New Mexico Consortium (NMC). 
  RA and FB also gratefully acknowledge support from the Center for Integrated Nanotechnologies (CINT) where these experiments were conducted.


\bibliographystyle{IEEEtran}   
\bibliography{bibliography}

\end{document}